\documentclass{article}
\usepackage{spconf,amsmath,graphicx}
\usepackage{graphics} 
\usepackage{epstopdf} 
\usepackage{mathptmx} 
\usepackage{times} 
\usepackage{amsmath} 
\usepackage{amssymb}  
\usepackage{cite}
\usepackage{multirow}
\usepackage{subcaption}
\usepackage{float}
\usepackage{algorithmicx}
\usepackage[ruled]{algorithm}
\usepackage{algpseudocode}
\usepackage{xcolor}
\usepackage{url}

\def\sqz{\vspace{-3pt}}
\let\OLDthebibliography\thebibliography
\renewcommand\thebibliography[1]{
  \OLDthebibliography{#1}
  \setlength{\parskip}{0pt}
  \setlength{\itemsep}{0pt plus 0.3ex}
}

\title{DEEP MR IMAGE SUPER-RESOLUTION USING STRUCTURAL PRIORS }
 \name{Venkateswararao Cherukuri$^{1, 2}$ \qquad Tiantong Guo$^{1}$ \qquad Steven J. Schiff $^{2, 3}$ \qquad Vishal Monga$^{1,2}$\thanks{This work is supported by NIH Grant R01HD085853.}}

 \address{\small{$^{1}$ Dept. of Electrical Engineering, $^{2}$ Center for Neural Engineering, $^{3}$ Dept. Neurosurgery, Engineering Science and Mechanics, and Physics,}\\
 \small{The Pennsylvania State University}\sqz\sqz\sqz\sqz\sqz\sqz}
\begin{document}
%
\maketitle

\begin{abstract}
High resolution magnetic resonance (MR) images are desired for accurate diagnostics. In practice, image resolution is restricted by factors like hardware, cost and processing constraints. Recently, deep learning methods have been shown to produce compelling state of the art results for image super-resolution. Paying particular attention to desired hi-resolution MR image structure, we propose a new regularized network that exploits image priors, namely a low-rank structure and a sharpness prior to enhance deep MR image superresolution.  Our contributions are then incorporating these priors in an analytically tractable fashion in the learning of a convolutional neural network (CNN) that accomplishes the super-resolution task. This is particularly challenging for the low rank prior, since the rank is not a differentiable function of the image matrix (and hence the network parameters), an issue we address by pursuing differentiable approximations of the rank. Sharpness is emphasized by the variance of the Laplacian which we show can be implemented by a fixed {\em feedback} layer at the output of the network. Experiments performed on two publicly available MR brain image databases exhibit promising results particularly when training imagery is limited.

\end{abstract}
\sqz
\begin{keywords}
Super Resolution, Deep Learning, MR Image Processing
\end{keywords}
\section{Introduction}
\label{sec:intro}
\vspace{-.3cm}
High resolution MR images can provide rich structural information about bodily organs which is critical in analyzing a given medical condition. Often, the quality of these images is restricted by factors like imaging hardware, sensor noise, cost and time constraints. In such scenarios, the spatial resolution of these images can be enhanced by a well-designed mathematical algorithm. Simple and fast interpolation methods like bilinear, bicubic \cite{lehmann1999survey} have been widely used for increasing the size of low resolution (LR) medical images. In many cases, these methods are known to introduce blurring, blocking artifacts, ringing and are thus unable to recover sharp details of an image. To alleviate this problem, an alternative approach known as super-resolution (SR) was introduced in \cite{tsai1984multiframe}. Current literature on SR can be classified into two categories: multi-image SR and single-image SR.\\
In multi-image SR \cite{tsai1984multiframe, farsiu2004fast}, a HR image is generated by exploiting the information from multiple LR images which are acquired from the same scene with a slightly shifted field of view. However, these methods are likely to fail if an adequate amount of LR images from the same scene are not available. As an alternative approach, single image SR was introduced wherein multiple LR images from the same scene are not required to obtain a HR image. In this approach, a mapping between LR and HR images is learned by constructing examples from a given database \cite{trinh2014novel, freeman2002example, chang2004super, yang2010image, bahrami2016reconstruction}. \\
Recently, deep learning methods have been shown to produce compelling state of the art results \cite{dong2016image,kim2016accurate,guo2017deep,wang2015deep,dong2016accelerating,timofte2017ntire,kim2016deeply} for image SR.  Invariably though, the training burden of deep networks, i.e. the number of example LR and HR images (or patches), is quite significant. In some medical diagnosis problems, generous LR and HR pairs is not a problem but there are compelling real-world problems such as enhancing 3T MR to 7T MR images \cite{bahrami2016reconstruction}, where the paucity of training has been recognized.  There has been encouraging recent application of deep networks for MR image SR \cite{yang2016super, srinivasan2017super} but the methods remain training intensive. An outstanding open challenge for deep MR image superresolution is the development of methods that exhibit a graceful degradation with respect to (w.r.t.) the number of training LR and HR image pairs. \\
\begin{figure}
	\begin{center}
		\includegraphics[scale=.045]{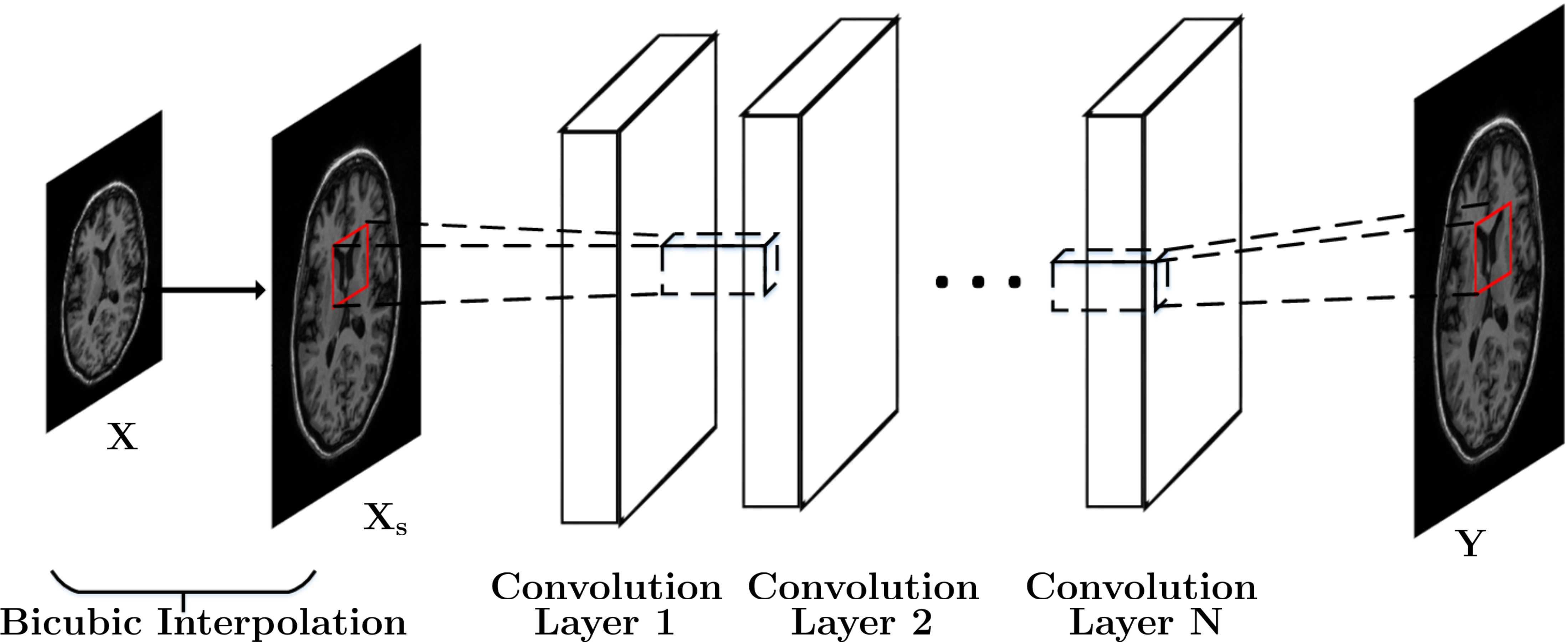}
	\end{center}
	\vspace{-.5cm}\sqz
	\caption{\small{SRCNN network.}}\sqz\sqz\sqz\sqz\sqz\sqz\sqz\sqz\sqz
	\label{fig:SRCNN}
\end{figure}
\textbf{Motivation and Contributions:} Our approach to improve deep MR image superresolution, even in the face of limited training is via the exploitation of suitable prior information pertinent to MR images. In \cite{shi2015lrtv}, a model based SR approach is presented that uses low-rank (approximated by nuclear norm) and total variation regularizers. The authors in \cite{shi2015lrtv} validate that MR images from various parts of the body can be reconstructed with a peak signal-to-noise ratio (PSNR) of close to $50$ db by retaining only half of the singular values obtained by a singular value decomposition (SVD) of the image matrix. Despite this promise, using a rank or even its nuclear norm relaxation in a deep network for SR presents a stiff analytical challenge since neither is a differentiable function of the image matrix (and hence the network parameters). Our contribution includes incorporating a suitable approximation to the rank, which is smooth, differentiable and amenable for learning in a deep CNN framework. Additionally, recognizing the need for sharp well formed edges in diagnosis, we propose a sharpness prior realized via a variance of the laplacian measure which adds to the network structure at the output as a fixed (non-optimizable) feedback layer.  We use a CNN for super-resolution (SRCNN) as described in \cite{dong2016image} as our base network. Our SR method is then called deep network with structural priors (DNSP).

\sqz\sqz
\section{Deep Learning for MR Image Super-resolution}
\label{sec:DNSP}
\vspace{-.3cm}
\subsection{Notation:}
Let $X \in \mathbb{R}^{M\times N}$ represent the LR image where $M$ and $N$ are the width and height of the image respectively. Let $Y\in \mathbb{R}^{sM\times sN}$ be the output HR image and $s$ is the desired scale to which $X$ needs to be upscaled and $Y_{g} \in \mathbb{R}^{sM\times sN}$ is the ground truth HR image for $X$. Let $W_{k}^{l} \in \mathbb{R}^{m\times n\times d}$ be the $k^{th}$ convolutional filter in layer $l$ where $m$, $n$ and $d$ represent the width, height and depth of the filter respectively. Similarly, let $b_{k}^{l} \in \mathbb{R}$ be the $k^{th}$ bias coefficient of layer $l$. The objective of the network is to learn $W_{k}^{l}$ and $b_{k}^{l}$ so that the output of the network $Y$ is a close representation of the ground truth $Y_{g}$. So, let $\Theta = \{W_{k}^{l}, b_{k}^{l}\} \forall l,k$. To make the size of input and output of the network the same, we first upscale $X$ by a factor of $s$ using bicubic interpolation and use this upscaled $X_{s}\in \mathbb{R}^{sM\times sN}$ as input to the network. Finally, let the mapping function of the network be represented by $F$ where $F(X_{s}, \Theta) = Y$.

\subsection{Deep CNNs For SR}
Deep learning methods are a class of machine learning methods which are inspired from biological neural networks. In general, a cascade of many nonlinear processing units are used to learn features to represent data effectively for a  given task. In particular, a deep CNN for image SR usually consists of two or more convolutional layers (each layer essentially is a combination of filters followed by an activation function) which are used to learn an end-to-end  mapping between sample HR and LR images. For example, Figure \ref{fig:SRCNN} illustrates the SRCNN network \cite{dong2016image, yang2016super} for super-resolution. Each convolutional layer in the network consists of several learnable filters, which are convolved with output from the previous layer. For a given layer, outputs obtained by convoluting with each filter are combined to form a data cube which is passed through a nonlinear activation function and then forwarded as an input to next layer \cite{lecun2015deep}. Most commonly used activation function in recent times is the Rectified linear unit (Relu) \cite{glorot2011deep}. The input to the first layer is the image obtained after bicubic interpolation and the output of the last layer is the expected HR image. The filters are learned to minimize the loss function given by:\sqz\sqz\sqz
\begin{equation}\label{eq1}
E(\Theta) = \frac{1}{2}\|Y_{g} - F(X_{s}, \Theta)\|_{F}^{2}\sqz\sqz\sqz
\end{equation}
where $\parallel\bullet\|_{F}$ represents the Frobenius norm.
\sqz\sqz
\section{Deep Networks with Structural Priors}
\sqz\sqz
As discussed in Section \ref{sec:intro}, we integrate two priors into the learning of the CNN. Note that both the priors are to be applied on $Y$ as it represents the desired output HR image. The two priors are as follows: \\
\textbf{Low Rank Prior:} It has been argued recently \cite{shi2015lrtv} that MR images are naturally rank deficient. However, the rank of a matrix is a non-differentiable function w.r.t. its input and therefore cannot be used as regularizer in a CNN. Most of the optimization problems with a low-rank constraint are solved by minimizing the nuclear norm of the matrix which is a convex relaxation for the low-rank constraint. However, this relaxation cannot be used in a CNN as the nuclear norm is also a non-differentiable function. Recently, a function which is smooth and differentiable was proposed in \cite{malek2014recovery} that can approximate the rank of a matrix accurately. It is defined as:\sqz\sqz
\begin{equation}\label{eq2}
G_{\delta}(Y) = h_{\delta}(\pmb{\sigma}(Y))\sqz\sqz
\end{equation}
where $h_{\delta}(\pmb{\sigma}(Y)) = \sum_{i=1}^{R}g_{\delta}(\sigma_{i}(Y))$, $\sigma_{i}(Y)$ represents the $i^{th}$ singular value of $Y$ and\sqz\sqz\sqz\sqz
\begin{equation}\label{eq3}
g_{\delta}(x) = \exp\Bigg(-\frac{x^{2}}{2\delta^{2}}\Bigg)\sqz\sqz
\end{equation}
where $\delta$ is a tunable parameter that affects the measure of approximation error in finding the rank\footnote{We chose $\delta = .01$ based on guidelines mentioned in \cite{malek2014recovery}.}. Intuitively $G_{\delta}(Y)$ gives the number of singular values of $Y$ which are zero. Therefore, $\mbox{rank}(Y)\approx R - G_{\delta}(Y)$. Let $R_{\delta}(Y) = R - G_{\delta}(Y)$, where $R = \min(sM, sN)$. Now, the function $R_{\delta}(Y)$ is differentiable and its gradient w.r.t. $Y$ is given by:
\begin{equation}\label{eq4}
-U\mbox{diag}\Bigg(-\frac{\sigma_{1}}{\delta^{2}}e^{-\sigma_{1}^{2}/2\delta^{2}},\ldots,-\frac{\sigma_{R}}{\delta^{2}}e^{-\sigma_{R}^{2}/2\delta^{2}}\Bigg)Z^{T}
\end{equation}
where SVD of $Y = U\mbox{diag}(\sigma_{1}, \ldots, \sigma_{R})Z^{T}$. \\ 
\begin{figure}
 \begin{center}
  \includegraphics[width=\linewidth]{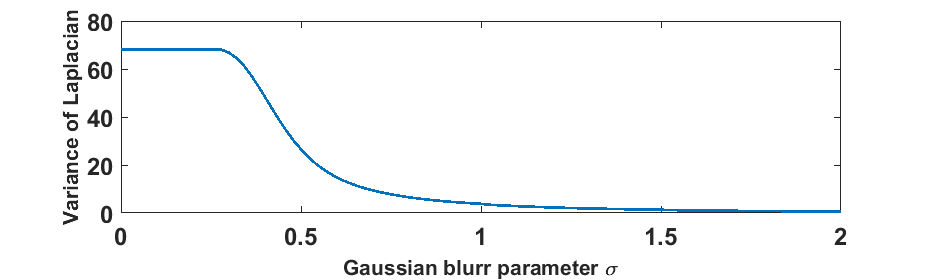}
 \end{center}
  \vspace{-.5cm}
  \caption{\small{Variance of the Laplacian vs increasing the blur parameter.}}\sqz\sqz\sqz\sqz\sqz\sqz
  \label{fig:sharpDemo}
\end{figure}
\begin{figure*}[t]
	\begin{center}
		\includegraphics[width=0.9\textwidth]{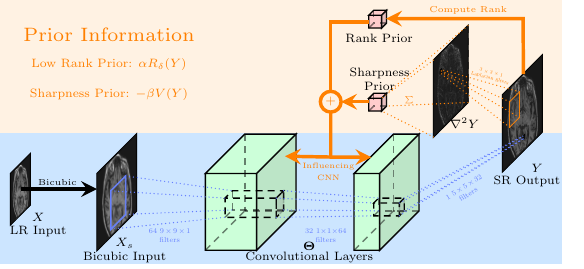}
	\end{center}
	\vspace{-.5cm}
	\caption{\small{Deep Network with Structural Priors (DNSP) for MR image super-resolution. Note the prior processing (shown in orange) is used {\em only} in the learning of the network. For a given test LR input image $X$, the learned CNN is used to generate the output SR image $Y$. }}\sqz\sqz\sqz\sqz
	\label{fig:Net}
\end{figure*}
\textbf{Sharpness Prior:} HR images look much sharper compared to LR images. The main reason can be attributed to blurriness of the LR images. The pursuit of quantifying sharpness begins by computing the Laplacian ($\nabla^{2}Y$) of the image \cite{forsyth2011computer}. The laplacian of a smooth/blurred image is more uniform compared to the laplacian of a sharp image. The variance of the Laplacian is hence an indicator of sharpness. As shown in Figure \ref{fig:sharpDemo}, an MR brain image is degraded by a gaussian filter with different blur parameters and plotted against the variance of laplacian. It can be observed that the variance of laplacian decreases as the blur parameter increases. Therefore, we propose to use $V(Y) = var(\nabla^{2}Y)$ as a regularizer to encourage the CNN to yield sharper HR images. $V(Y)$ is a quadratic function in Y and therefore a differentiable function which can be easily integrated into the CNN learning. Note that the laplacian of an image can be implemented by well-known linear filters \cite{forsyth2011computer}, which are also easily integrated into the CNN via a filtering layer at the output as shown in Fig. \ref{fig:Net}.

\noindent \textbf{Network Structure:} We incorporate the aforementioned two priors into the the basic SRCNN \cite{dong2016image} framework and the proposed Deep Network with Structural Priors (DNSP) is shown in Fig. \ref{fig:Net}. Note that our priors can be integrated into any other deep SR network as well. There are 3 layers in DNSP: the first layer has $64$ $9\times 9\times 1$ filters, the second layer has $32$ $1\times 1\times 64$ filters while the final layer has one $5\times 5\times 32$ filter. The output of each layer except that of the final layer is fed into ReLU to generate a nonlinear activation map \cite{nair2010rectified}. We also use a $3\times 3$ filter $L = [[0 \mbox{ }- 1 \mbox{ }0]^{T} [-1 \mbox{   4 } \mbox{ }-1]^{T} [0\mbox{ } - 1\mbox{ } 0]^{T}]$ after the final layer to compute the Laplacian and subsequently find the variance of Laplacian. Finally, the loss function of DNSP to be minimized is given by:\vspace{-5pt}
\begin{equation}\label{eq5}
E(\Theta) = \underbrace{\frac{1}{2}\|Y_{g} - F(X_{s}, \Theta)\|_{F}^{2}}_{MSE} + \underbrace{\alpha R_{\delta}(Y)}_{Low Rank} - \underbrace{\beta V(Y)}_{Sharpness Prior}\sqz\sqz
\end{equation}
where, $Y = F(X_{s}, \Theta)$, $\alpha$ and $\beta$ are positive regularization parameters\footnote{We chose $\alpha = .1$ and $\beta = 5\times 10^{-5}$ by cross validation.}, note that negative sign before $V(Y)$ is to increase the variance of Laplacian. Note that the SRCNN loss function in Eq (\ref{eq1}) is a special case of Eq. (\ref{eq5}). We learn $\Theta$ by minimizing $E(\Theta)$ using a stochastic gradient descent method \cite{lecun1998gradient, werbos1994roots}. In particular, weights are updated by the following equation:\sqz\sqz
\begin{equation}\label{eqUpdate}
\Theta^{t+1} = \Theta^{t} - \eta\frac{\partial E}{\partial\Theta^{t}}\sqz\sqz
\end{equation}
where, $t$ represents the iteration number, $\eta$ represents the learning rate, and $\Theta^{t}$ represents the values of weights at previous iteration. As $\Theta = \{W_{k}^{l}, b_{k}^{l}\} \forall l,k$, following gradients are to be computed: $\frac{\partial E}{\partial w_{k}^{l}}$, $\frac{\partial E}{\partial b_{k}^{l}}$, where $w_{k}^{l}$ denotes an arbitrary scalar entry in filter $W_{k}^{l}$. For simplicity, let output image $Y$ be of dimension $N\times N$. The equation for computing the gradient of weight $w_{k}^{l}$ in layer $l\in\{1,2,3\}$ is given by:\vspace{-5pt}
\begin{equation}\label{eqW3}
\frac{\partial E}{\partial w_{k}^{l}} = -(Y_{g} - Y)\diamond\frac{\partial Y}{\partial w_{k}^{l}} + \alpha D_{R_{\delta}}\diamond\frac{\partial Y}{\partial w_{k}^{l}} - \beta D_{V}\diamond\frac{\partial Y}{\partial w_{k}^{l}}\vspace{-2pt}
\end{equation}
where $\diamond$ between two matrices $A$ and $B$ is defined as $\sum_{i,j}A_{i,j}B_{i,j}$, $D_{R_{\delta}} = -U\mbox{diag}\Bigg(-\frac{\sigma_{1}}{\delta^{2}}e^{-\sigma_{1}^{2}/2\delta^{2}},\ldots,-\frac{\sigma_{R}}{\delta^{2}}e^{-\sigma_{R}^{2}/2\delta^{2}}\Bigg)Z^{T}$ is the gradient of $R_{\delta}(Y)$ and $D_{V}$ is the gradient for $V(Y)$. The complete expression for $D_{V}$ is given by:\sqz\sqz
\begin{equation*}\resizebox{\linewidth}{!}{$
D_{V} = [v_{i,j}], \mbox{  }v_{i,j} = d_{i,j} - \frac{1}{4}(d_{i-1,j} + d_{i+1,j} + d_{i, j-1} + d_{i, j+1})$},\sqz
\end{equation*}
\begin{equation*}\resizebox{\linewidth}{!}{$
d_{i,j} = \frac{2}{(N^{2})(N^{2} - 1)}\big(N^{2}p_{i,j} - \sum_{a}\sum_{b}p_{a,b} - \sum_{m}\sum_{n}(p_{m,n} - \frac{\sum_{a}\sum_{b}p_{a,b}}{N^{2}})\big)$},\sqz
\end{equation*}
where $P = [p_{i,j}]$, and $P$ is obtained by convolving $Y$ with a $3\times 3$ laplacian operator $L$. Expression for $p_{i,j}$ is given by:\sqz\sqz
\begin{equation*}\resizebox{\linewidth}{!}{$\label{eqDv3}
p_{i,j} = y_{i,j} - \frac{1}{4}(y_{i-1,j} + y_{i+1,j} + y_{i, j-1} + y_{i, j+1}), \mbox{ and } Y = [y_{i,j}]
$}\sqz\sqz\end{equation*}
Detailed derivations for the above equations are reported in a technical report at \cite{PSU_venkat}. Note that the gradient for bias terms are also updated in a similar fashion. The partial derivative $\frac{\partial Y}{\partial {w_{k}}^{l}}$ is obtained by a standard back propagation rule \cite{lecun1998gradient, werbos1994roots}.

\section{Experimental Evaluation}
\sqz
\label{sec:Results}
\textbf{Databases:} We evaluate the proposed DNSP on two publicly available MR brain image databases. The first database is 20 simulated T1 brain image stacks from Brainweb (BW)\footnote{\url{http://brainweb.bic.mni.mcgill.ca/brainweb/}}. Axial slices of these 20 stacks are distributed evenly for training and evaluation purposes. From each stack, we extract 40 slices making a total of 400 images for training and 400 images for evaluation. The second database we work with is from the Alzheimer's Disease Neuroimaging Initiative (ADNI)\footnote{\url{http://adni.loni.usc.edu/}}. The training and evaluation configuration for this database is same as that of the BW database.\\
\textbf{LR image simulation: } Consistent with \cite{shi2015lrtv}, we simulate training LR images by applying a gaussian blur and factor of $2$ downsampling. These LR images are then upscaled by bicubic interpolation. To speed up the training process, we further extract patches of size $40\times 40$ from these bicubic enlarged LR training images. Note that this is also a standard procedure used for training a typical deep SR network \cite{dong2016image, kim2016accurate,guo2017deep}.\\
\textbf{Methods and Metrics for Comparison: }Two standard metrics PSNR and structural similarity index (SSIM) \cite{brunet2017optimizing} are used for evaluation. We compare  against: 1.) Bicubic interpolation (Bb), 2.) a competitive model based approach with low-rank and total variation (LRTV) regularizers \cite{shi2015lrtv}, 3.) example based super-res via sparse weighting (SRSW) \cite{trinh2014novel} -- a state-of-the art sparsity based method and 4.) SRCNN \cite{dong2016image, yang2016super} that is the most popular embodiment of a deep SR network.\\

\vspace{-.5cm}
\begin{table}[h]
\caption{\small{PSNR and SSIM comparisons}}
\label{tab:results}
\vspace{-.5cm}
\begin{center}
\resizebox{.6\linewidth}{!}{\begin{tabular}{cccc}
\hline\hline\ninept
\textbf{Method} & \textbf{Database} & \textbf{PSNR} & \textbf{SSIM}\\
\hline
\multirow{2}{*}{Bb} & BW & $29.09$ & $.8369$ \\
                             & ADNI  & $27.82$ & $.8958$ \\
\hline
\multirow{2}{*}{SRSW} & BW & $31.16$ & $.50$ \\
                             & ADNI  & $30.19$ & $.77$ \\
\hline
\multirow{2}{*}{LRTV} & BW & $30.46$ & $.856$ \\
                              & ADNI  & $30.50$ & $.783$ \\
\hline
\multirow{2}{*}{SRCNN} & BW &$32.37$ & $.8762$ \\
                              & ADNI  & $30.75$ & $.938$ \\
\hline
\multirow{2}{*}{DNSP} & BW &$\textbf{32.76}$ & $\textbf{.8788}$ \\
                              & ADNI  & $\textbf{31.27}$ & $\textbf{.9458}$ \\
\hline\hline
\end{tabular}}
\end{center}
\end{table}

\begin{figure}
 \begin{center}
  \includegraphics[scale=.22]{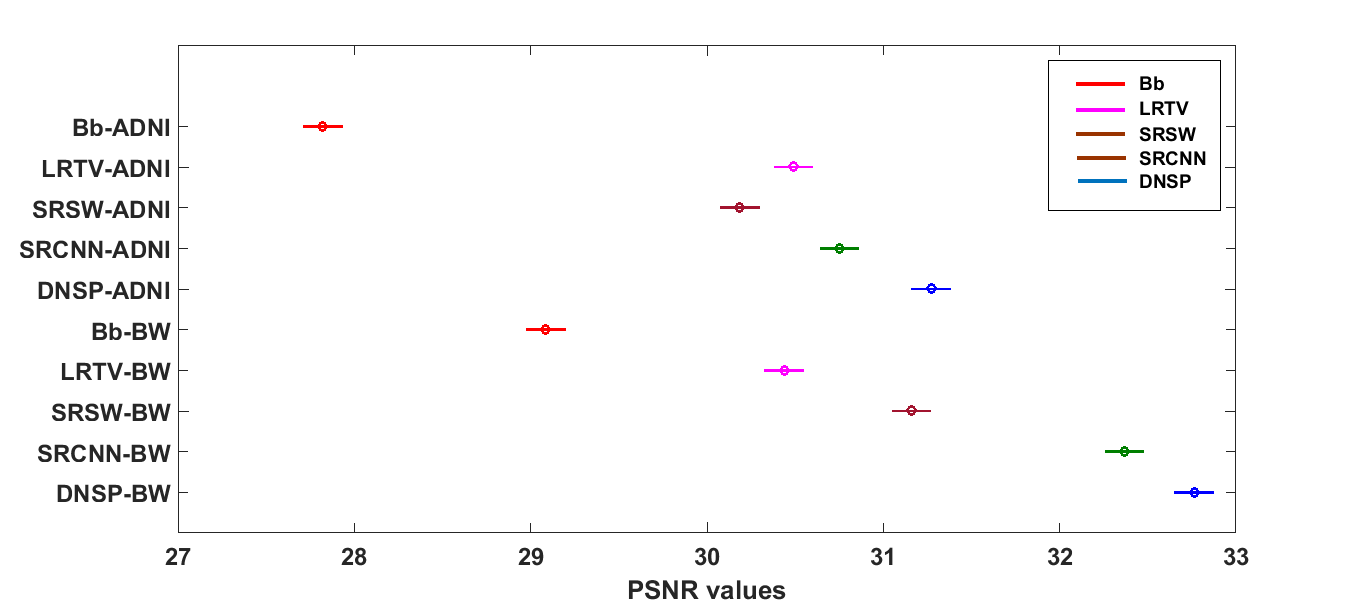}
 \end{center}
  \vspace{-.5cm}
  \caption{\small{2-way ANOVA comparing DNSP vs. competing methods. The intervals represent the 95 $\%$ confidence intervals of PSNR values for a given configuration of method-dataset. Values reported for ANOVA across the method factor are $df = 4$, $F = 1496.93$, $p\ll .01$.}}\vspace{-10pt}
  \label{fig:ANOVA}
\end{figure}

\begin{figure}
 \begin{center}
  \includegraphics[scale=.18]{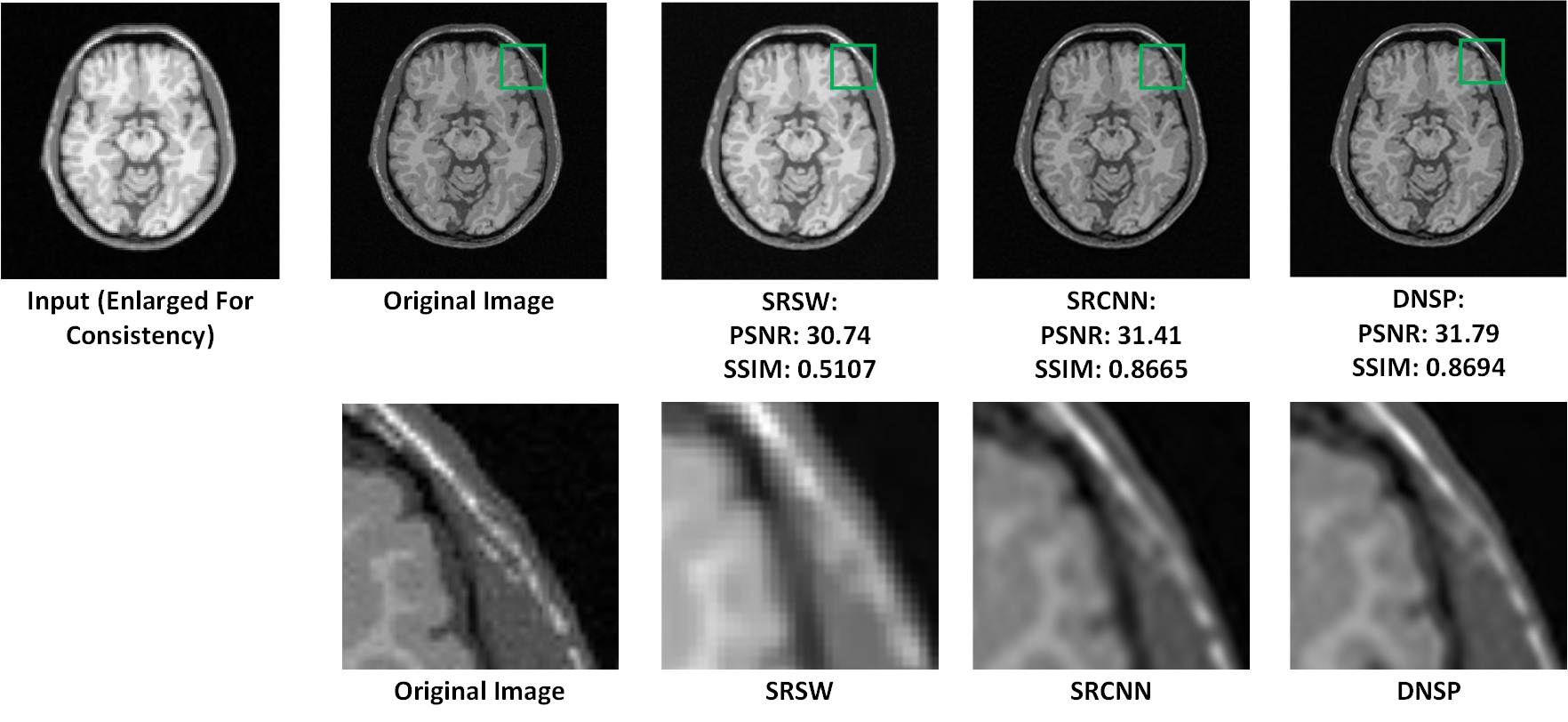}
 \end{center}
  \vspace{-.5cm}
  \caption{\small{Comparisons of top 3 methods w.r.t to PSNR for an image in BW data set. A small portion of the images (marked by green rectangle) in the first row is zoomed in and shown in second row.}}  \vspace{-15pt}
  \label{fig:images}
\end{figure}

\begin{figure}[h]
	\begin{center}
		\includegraphics[width=0.9\linewidth]{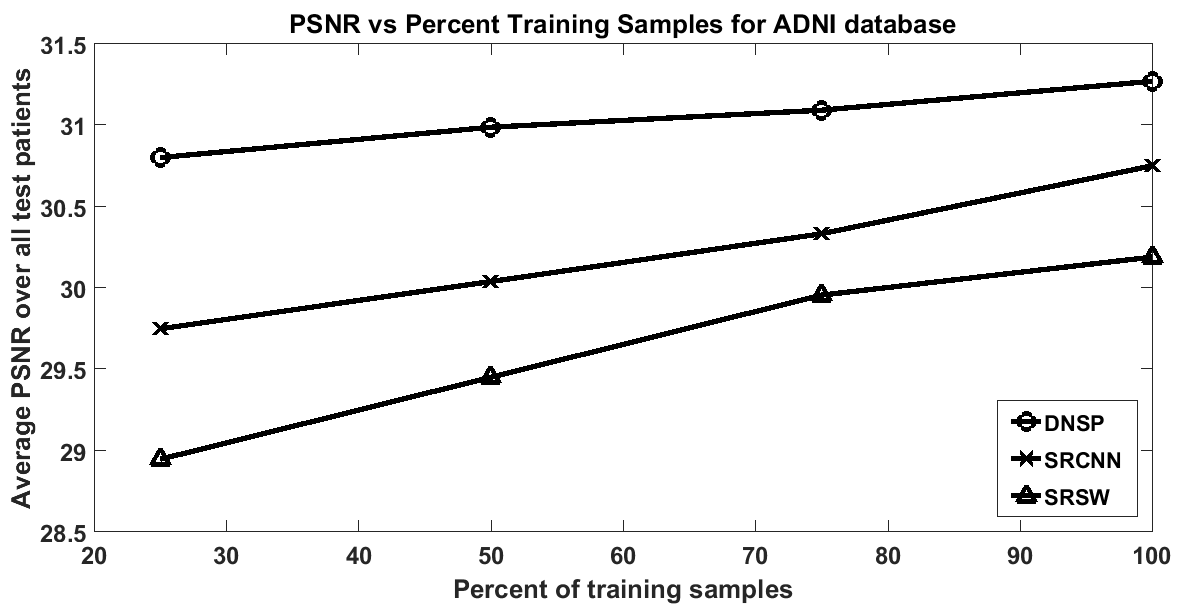}
	\end{center}
	\vspace{-.5cm}
	\caption{\small{PSNR vs percent training samples.}}
	\label{fig:trainPlot}
\end{figure}

\sqz\sqz\sqz\sqz\sqz\sqz
\noindent Table \ref{tab:results} shows PSNR and SSIM values for all competing methods. Two trends emerge from the results 1) DNSP outperforms the competition, and 2) Overall, deep SR methods, i.e.\ SRCNN and DNSP perform better. To confirm this statistically, we performed a 2-way Analysis of variance (ANOVA) on PSNR values for all the methods across the two datasets which is illustrated in Fig. \ref{fig:ANOVA}. It may be inferred from Fig. \ref{fig:ANOVA} that deep learning methods are statistically well separated from the traditional methods and further DNSP is well separated from SRCNN indicating the effectiveness of using prior information. Figure \ref{fig:images} illustrates the results of top 3 methods w.r.t. PSNR on a sample image from the BW database. DNSP performs better in recovering  finer image detail. \\
Figure \ref{fig:trainPlot} compares the performance of the learning based methods for different percentage of training samples considered on the ADNI dataset. Twenty five, $50$ and $75$ percent of the 400 training images are employed. Two inferences can be made: 1) DNSP consistently outperforms  SRCNN and SRSW, 2) The performance degradation of DNSP is more graceful. For example, PSNR values for SRCNN and SRSW dropped by almost close to 1db whereas for DNSP, the drop is around .5db, when the training drops to 25 percent.
Note that LRTV is excluded for this comparison since this is model based and not an example/learning based technique \cite{shi2015lrtv}.

\vspace{-.5cm}\sqz\sqz
\section{Conclusion}
\label{sec:Conclusion}
\sqz\sqz
We present a novel regularized deep network structure for MR image superresolution, which excels in varying training regimes. This is accomplished by using two structural priors on the expected output HR image: 1) a low-rank prior, and 2) a sharpness prior. While we demonstrate improvements by employing SRCNN \cite{dong2016image} as our base network, our proposal is versatile and the proposed priors can be applied to extend other deep SR networks \cite{kim2016accurate, wang2015deep,timofte2017ntire,kim2016deeply} as well.

\bibliographystyle{IEEEbib}
\ninept
\bibliography{refs}

\end{document}